# Combining Evaluation Metrics via the Unanimous Improvement Ratio and its Application to Clustering Tasks

**Enrique Amigó**                                    ENRIQUE@LSI.UNED.ES
**Julio Gonzalo**                                    JULIO@LSI.UNED.ES
*UNED NLP & IR Group, Juan del Rosal 16*
*Madrid 28040, Spain*

**Javier Artiles**                                   JAVIER.ARTILES@QC.CUNY.EDU
*Blender Lab, Queens College (CUNY),*
*65-30 Kissena Blvd, NY 11367, USA*

**Felisa Verdejo**                                   FELISA@LSI.UNED.ES
*UNED NLP & IR Group, Juan del Rosal 16*
*Madrid 28040, Spain*

## Abstract

Many Artificial Intelligence tasks cannot be evaluated with a single quality criterion and some sort of weighted combination is needed to provide system rankings. A problem of weighted combination measures is that slight changes in the relative weights may produce substantial changes in the system rankings. This paper introduces the *Unanimous Improvement Ratio* (UIR), a measure that complements standard metric combination criteria (such as van Rijsbergen's F-measure) and indicates how robust the measured differences are to changes in the relative weights of the individual metrics. UIR is meant to elucidate whether a perceived difference between two systems is an artifact of how individual metrics are weighted.

Besides discussing the theoretical foundations of UIR, this paper presents empirical results that confirm the validity and usefulness of the metric for the Text Clustering problem, where there is a tradeoff between precision and recall based metrics and results are particularly sensitive to the weighting scheme used to combine them. Remarkably, our experiments show that UIR can be used as a predictor of how well differences between systems measured on a given test bed will also hold in a different test bed.

## 1. Introduction

Many Artificial Intelligence tasks cannot be evaluated with a single quality criterion, and some sort of weighted combination is needed to provide system rankings. Many problems, for instance, require considering both Precision (P) and Recall (R) to compare systems' performance. Perhaps the most common combining function is the F-measure (van Rijsbergen, 1974), which includes a parameter $\alpha$ that sets the relative weight of metrics; when $\alpha = 0.5$, both metrics have the same relative weight and F computes their harmonic mean.

A problem of weighted combination measures is that relative weights are established intuitively for a given task, but at the same time a slight change in the relative weights may produce substantial changes in the system rankings. The reason for this behavior is that an overall improvement in F often derives from an improvement in one of the individual metrics at the expense of a decrement in the other. For instance, if a system A improves a system B in precision with a loss in recall, F may say that A is better than B or viceversa, depending on the relative weight of precision and recall (i.e. the $\alpha$ value).

This situation is more common than one might expect. Table 1 shows evaluation results for different tasks extracted from the ACL 2009 (Su et al., 2009) conference proceedings, in which P and R are combined using the F-measure. For each paper we have considered three evaluation results: the one that maximizes F, which is presented as the best result in the paper, the baseline, and an alternative method that is also considered. Note that in all cases, the top ranked system improves the baseline according to the F-measure, but at the cost of decreasing one of the metrics. For instance, in the case of the paper on Word Alignment, the average R grows from 54.82 to 72.49, while P decreases from 72.76 to 69.19. In the paper on Sentiment Analysis, P increases in four points but R decreases in five points. How reasonable is to assume that the contrastive system is indeed improving the baseline?

The evaluation results for the alternative approach are also controversial: in all cases, the alternative approach improves the best system according to one metric, and it is improved according to the other. Therefore, depending on the relative metric weighting, the alternative approach could be considered better or worse than the best scored system.

The conclusion is that the $\alpha$ parameter is crucial when comparing real systems. In practice, however, most authors set $\alpha = 0.5$ (equal weights for precision and recall) as a standard, agnostic choice that requires no further justification. Thus, without a notion of how much a perceived difference between systems depends on the relative weights between metrics, the interpretation of results with F – or any other combination scheme – can be misleading.

Our goal is, therefore, to find a way of estimating to what extent a perceived difference using a metric combination scheme – F or any other – is robust to changes in the relative weights assigned to each individual metric.

In this paper we propose a novel measure, the Unanimity Improvement Ratio (UIR), which relies on a simple observation: when a system A improves other system B according to all individual metrics (the improvement is *unanimous*), A is better than B for any weighting scheme. Given a test collection with $n$ test cases, the more test cases where improvements are unanimous, the more robust the perceived difference (average difference in F or any other combination scheme) will be.

In other words, as well as statistical significance tests provide information about the robustness of the evaluation across test cases (*Is the perceived difference between two systems an artifact of the set of test cases used in the test collection?*), UIR is meant to provide information about the robustness of the evaluation across variations of the relative metric weightings (*Is the perceived difference between two systems an artifact of the relative metric weighting chosen in the evaluation metric?*).

Our experiments on clustering test collections show that UIR contributes to the analysis of evaluation results in two ways:

- It allows to detect system improvements that are biased by the metric weighting scheme. In such cases, experimenters should carefully justify a particular choice of relative weights and check whether results are swapped in their vicinity.

| Systems | | Precision | Recall | F |
|---|---|---|---|---|
| Task: Word alignment (Huang 2009) | | | | |
| Baseline | BM | **72.76** | 54.82 | 62.53 |
| Max. F | Link-Select | 69.19 | **72.49** | **70.81** |
| Alternative | ME | 72.66 | 66.17 | 69.26 |
| Task: Opinion Question Answering (Li et al. 2009) | | | | |
| Baseline | System 3 | **10.9** | 21.6 | 17.2 |
| Max. F | OpHit | 10.2 | **25.6** | **20.5** |
| Alternative | OpPageRank | 10.9 | 24.2 | 20 |
| Task: Sentiment Analysis (Kim et al 2009) | | | | |
| Baseline | BASELINE | 30.5 | **86.6** | 45.1 |
| Max. F | VS-LSA-DTP | **34.9** | 71.9 | **47** |
| Alternative | VS PMI | 31.1 | 83.3 | 45.3 |
| Task: Lexical Reference Rule Extraction (Shnarch et al 2009) | | | | |
| Baseline | No expansion | **54** | **19** | 28 |
| Max. F | Wordnet+Wiki | 47 | **35** | **40** |
| Alternative | All rules + Dice filter | 49 | 31 | 38 |

Table 1: A few three-way system comparisons taken from ACL 2009 Conference Proceedings (Su et al., 2009)

- It increases substantially the consistency of evaluation results across datasets: a result that is supported by a high Unanimous Improvement Ratio is much more likely to hold in a different test collection. This is, perhaps, the most relevant practical application of UIR: as a predictor of how much a result can be replicable across test collections.

Although most of the work presented in this paper applies to other research areas, here we will focus on the clustering task as one of the most relevant examples because clustering tasks are specially sensitive to the metric relative weightings. Our research goals are:

1. To investigate empirically whether clustering evaluation can be biased by precision and recall relative weights in F. We will use one of the most recent test collections focused on a text clustering problem (Artiles, Gonzalo, & Sekine, 2009).

2. To introduce a measure that quantifies the robustness of evaluation results across metric combining criteria, which leads us to propose the UIR measure, which is derived from the Conjoint Measurement Theory (Luce & Tukey, 1964).

3. To analyze empirically how UIR and F-measure complement each other.

4. To illustrate the application of UIR when comparing systems in the context of a shared task, and measure how UIR serves as a predictor of the consistency of evaluation results across different test collections.

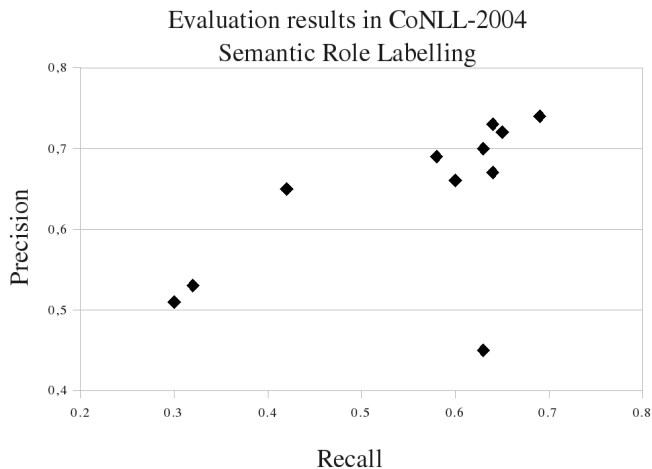

Figure 1: Evaluation results for semantic labeling in CoNLL 2004

## 2. Combining Functions for Evaluation Metrics

In this section we briefly review different metrics combination criteria. We present the rationale behind each metric weighting approach as well as its effects on the systems ranking.

### 2.1 F-measure

The most frequent way of combining two evaluation metrics is the *F-measure* (van Rijsbergen, 1974). It was originally proposed for the evaluation of Information Retrieval systems (IR), but its use has expanded to many other tasks. Given two metrics P and R (e.g. precision and recall, Purity and Inverse Purity, etc.), van Rijsbergen's F-measure combines them into a single measure of efficiency as follows:

$$F(R,P) = \frac{1}{\alpha(\frac{1}{P}) + (1-\alpha)(\frac{1}{R})}$$

F assumes that the $\alpha$ value is set for a particular evaluation scenario. This parameter represents the relative weight of metrics. In some cases the $\alpha$ value is not crucial; in particular, when metrics are correlated. For instance, Figure 1 shows the precision and recall levels obtained at the CoNLL-2004 shared task for evaluating Semantic Role Labeling systems (Carreras & Màrquez, 2004). Except for one system, every substantial improvement in precision involves also an increase in recall. In this case, the relative metric weighting does not substantially modify the system ranking.

In cases where the metrics are not completely correlated, the *Decreasing Marginal Effectiveness* property (van Rijsbergen, 1974) ensures a certain robustness across $\alpha$ values. F satisfies this property, which states that a large decrease in one metric cannot be compensated by a large increase in the other metric. Therefore, systems with very low precision or recall will obtain low F-values for any $\alpha$ value. This is discussed in more detail in Section 5.1. However, as we will show in Section 3.4, in other cases the *Decreasing Marginal*

*Effectiveness* property does not prevent the F-measure from being overly sensitive to small changes in the $\alpha$ value.

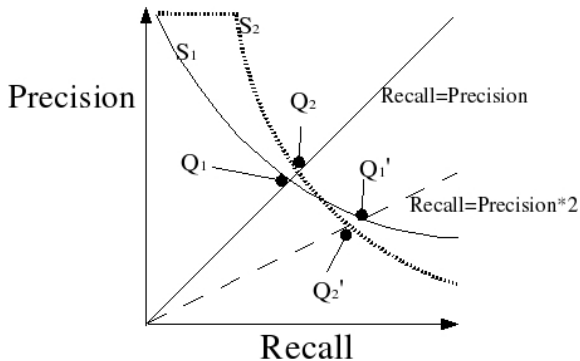

Figure 2: An example of two systems evaluated by break-even point

## 2.2 Precision and Recall Break-even Point

Another way of combining metrics consists of evaluating the system at the point where one metric equals the other (Tao Li & Zhu, 2006). This method is applicable when each system is represented by a trade-off between both metrics, for instance, a precision/recall curve. This method relies on the idea that increasing both metrics implies necessarily a overall quality increase. For instance, it assumes that obtaining a 0.4 of precision at the recall point 0.4 is better than obtaining a 0.3 of precision at the recall point 0.3.

Actually, the break-even point assumes the same relevance for both metrics. It considers the precision/recall point where the system distributes its efforts equitably between both metrics. Indeed, we could change the relative relevance of metrics when computing the break-even point.

Figure 2 illustrates this idea. The continuous curve represents the trade-off between precision and recall for system $S1$. The straight diagonal represents the points where both metrics return the same score. The quality of the system corresponds therefore with the intersection between this diagonal and the precision/recall curve. On the other hand, the discontinuous curve represents another system $S2$ which achieves an increase of precision at low recall levels at the cost of decreasing precision at high recall levels. According to the break-even points, the second system is superior than the first one.

However, we could give more relevance to recall identifying the point where recall doubles precision. In that case, we would obtain the intersection points $Q_1'$ and $Q_2'$ shown in the figure, which reverses the quality order between systems. In conclusion, the break-even point also assumes an arbitrary relative relevance for the combined metrics.

## 2.3 Area Under the Precision/Recall Curve

Some approaches average scores over every potential parameterization of the metric combining function. For instance, Mean Average Precision (MAP) is oriented to IR systems, and computes the average precision across a number of recall levels. Another example is the Receiver Operating Characteristic (ROC) function used to evaluate binary classifiers (Cormack & Lynam, 2005). ROC computes the probability that a positive sample receives a confidence score higher than a negative sample, independently from the threshold used to classify the samples. Both functions are related with the area AUC that exists under the precision/recall curve (Cormack & Lynam, 2005).

In both MAP and ROC the low and high recall regions have the same relative relevance when computing this area. Again, we could change the measures in order to assign different weights to high and low recall levels. Indeed in (Weng & Poon, 2008) a weighted Area Under the Curve is proposed. Something similar would happen if we average F across different $\alpha$ values.

Note that these measures can only be applied in certain kinds of problem, such as binary classification or document retrieval, where the system output can be seen as a ranking, and different cutoff points in the ranking give different precision/recall values. They are not directly applicable, in particular, to the clustering problem which is the focus of our work here.

## 3. Combining Metrics in Clustering Tasks

In this section we present metric combination experiments on a specific clustering task. Our results corroborate the importance of quantifying the robustness of systems across different weighting schemes.

### 3.1 The Clustering Task

Clustering (grouping similar items) has applications in a wide range of Artificial Intelligence problems. In particular, in the context of textual information access, clustering algorithms are employed for Information Retrieval (clustering text documents according to their content similarity), document summarization (grouping pieces of text in order to detect redundant information), topic tracking, opinion mining (e.g. grouping opinions about a specific topic), etc.

In such scenarios, clustering distributions produced by systems are usually evaluated according to their similarity to a manually produced gold standard (*extrinsic* evaluation). There is a wide set of metrics that measure this similarity (Amigó, Gonzalo, Artiles, & Verdejo, 2008), but all of them rely on two quality dimensions: (i) to what extent items in the same cluster also belong to the same group in the gold standard; and (ii) to what extent items in different clusters also belong to different groups in the gold standard. A wide set of extrinsic metrics has been proposed: Entropy and Class Entropy (Steinbach, Karypis, & Kumar, 2000; Ghosh, 2003), Purity and Inverse Purity (Zhao & Karypis, 2001), precision and recall Bcubed metrics (Bagga & Baldwin, 1998), metrics based on counting pairs (Halkidi, Batistakis, & Vazirgiannis, 2001; Meila, 2003), etc.[1]

---

1. See the work of Amigó et al. (2008) for a detailed overview.

## 3.2 Dataset

WePS (Web People Search) campaigns are focused on the task of disambiguating person names in Web search results. The input for systems is a ranked list of web pages retrieved from a Web search engine using a person name as a query (e.g. "John Smith"). The challenge is to correctly estimate the number of different people sharing the name in the search results and group documents referring to the same individual. For every person name, WePS datasets provide around 100 web pages from the top search results, using the quoted person name as query. In order to provide different ambiguity scenarios, person names were sampled from the US Census, Wikipedia, and listings of Program Committee members of Computer Science Conferences.

Systems are evaluated comparing their output with a gold standard: a manual grouping of documents produced by two human judges in two rounds (first they annotated the corpus independently and then they discussed the disagreements together). Note that a single document can be assigned to more than one cluster: an Amazon search results list, for instance, may refer to books written by different authors with the same name. The WePS task is, therefore, an *overlapping* clustering problem, a more general case of clustering where items are not restricted to belong to one single cluster. Both the WePS datasets and the official evaluation metrics reflect this fact.

For our experiments we have focused on the evaluation results obtained in the WePS-1 (Artiles, Gonzalo, & Sekine, 2007) and WePS-2 (Artiles et al., 2009) evaluation campaigns. The WePS-1 corpus also includes data from the Web03 test bed (Mann, 2006), which was used for trial purposes and follows similar annotation guidelines, although the number of document per ambiguous name is more variable. We will refer to these corpora as WePS-1a (trial), WePS-1b and WePS-2 [2].

## 3.3 Thresholds and Stopping Criteria

The clustering task involves three main aspects that determine the system's output quality. The first one is the method used for measuring similarity between documents; the second is the clustering algorithm (k-neighbors, Hierarchical Agglomerative Clustering, etc.); and the third aspect to be considered usually consists of a couple of related variables to be fixed: a similarity threshold – above which two pages will be considered as related – and a stopping criterion which determines when the clustering process stops and, consequently, the number of clusters produced by the system.

Figure 3 shows how Purity and Inverse Purity values change for different clustering stopping points, for one of the systems evaluated on the WePS-1b corpus [3]. Purity focuses on the frequency of the most common category into each cluster (Amigó et al., 2008). Being $C$ the set of clusters to be evaluated, $L$ the set of categories (reference distribution) and $N$ the number of clustered items, Purity is computed by taking the weighted average of maximal precision values:

$$\text{Purity} = \sum_i \frac{|C_i|}{N} \max_j \text{Precision}(C_i, L_j)$$

where the precision of a cluster $C_i$ for a given category $L_j$ is defined as:

$$\text{Precision}(C_i, L_j) = \frac{|C_i \bigcap L_j|}{|C_i|}$$

---

2. The WePS datasets were selected for our experiments because (i) they address a relevant and well-defined clustering task; (ii) its use is widespread: WePS datasets have been used in hundreds of experiments since the first WePS evaluation in 2007; (iii) runs submitted by participants to WePS-1 and WePS-2 were available to us, which was essential to experiment with different evaluation measures. WePS datasets are freely available from http://nlp.uned.es/weps.

3. This system is based on bag of words, TF/IDF word weighting, stopword removal, cosine distance and a Hierarchical Agglomerative Clustering algorithm.

Purity penalizes the noise in a cluster, but it does not reward grouping items from the same category together; if we simply make one cluster per item, we reach trivially a maximum purity value. Inverse Purity focuses on the cluster with maximum recall for each category. Inverse Purity is defined as:

$$\text{Inverse Purity} = \sum_i \frac{|L_i|}{N} \max_j \text{Precision}(L_i, C_j)$$

Inverse Purity rewards grouping items together, but it does not penalize mixing items from different categories; we can reach a maximum value for Inverse purity by making a single cluster with all items.

Any change in the stopping point implies an increase in Purity at the cost of a decrease in Inverse Purity, or viceversa. Therefore, each possible $\alpha$ value in F rewards different stopping points. This phenomenon produces a high dependency between clustering evaluation results and the metric combining function.

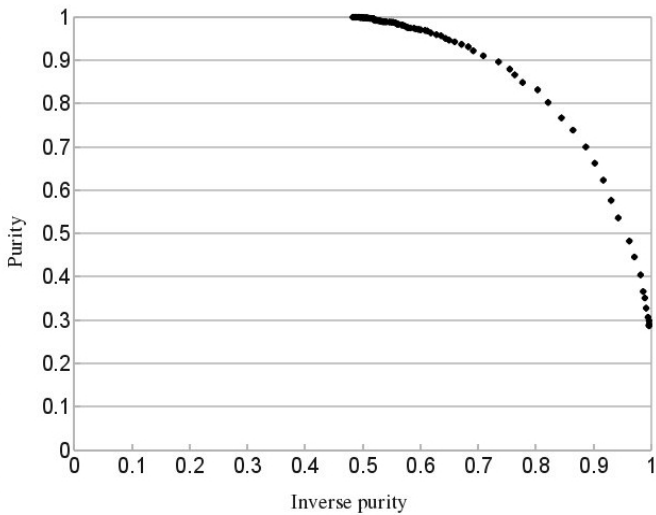

Figure 3: An example of the trade-off between Purity and Inverse Purity when optimizing the grouping threshold

### 3.4 Robustness Across $\alpha$ Values

Determining the appropriate $\alpha$ value for a given scenario is not trivial. For instance, from a user's point of view in the WePS task, it is easier to discard a few irrelevant documents from the good cluster (because its precision is not perfect but it has a high recall) than having to check for additional relevant documents in all clusters (because its precision is high but its recall is not). Therefore, it seems that Inverse Purity should have priority over Purity, i.e., the value of $\alpha$ should be below 0.5. From the point of view of a company providing a web people search service, however, the situation is quite different: their priority is having a very high precision, because mixing the profiles of, say, a criminal and a doctor may result in the company being sued. From their perspective, $\alpha$ should receive a high value. The WePS campaign decided to be agnostic and set a neutral $\alpha = 0.5$ value.

Table 2 shows the resulting system ranking in WePS-1b according to F with $\alpha$ set at 0.5 and 0.2. This ranking includes two baseline systems: $B_1$ consists of grouping each document in a separate cluster, and $B_{100}$ consists of grouping all documents into one single cluster. $B_1$ maximizes Purity, and $B_{100}$ maximizes Inverse Purity.

$B_1$ and $B_{100}$ may obtain a high or low F-measure depending on the $\alpha$ value. As the table shows, for $\alpha = 0.5$ $B_1$ outperforms $B_{100}$ and also a considerable number of systems. The reason for this result is that, in the WePS-1b test set, there are many singleton clusters (people which are referred to in only one web page). This means that a default strategy of making one cluster per document will not only achieve maximal Purity, but also an acceptable Inverse Purity (0.45). However, if $\alpha$ is fixed at 0.2, $B_1$ goes down to the bottom of the ranking and it is outperformed by all systems, including the other baseline $B_{100}$.

Note that outperforming a trivial baseline system such as $B_1$ is crucial to optimize systems, given that the optimization cycle could otherwise lead to a baseline approach like $B_1$. The drawback of $B_1$ is that it is not informative (the output does not depend on the input) and, crucially, it is very sensitive to variations in $\alpha$. In other words, its performance is not robust to changes in the metric combination criterion. Remarkably, the top scoring system, $S_1$, is the best for both $\alpha$ values. Our primary motivation in this article is to quantify the robustness across $\alpha$ values in order to complement the information given by traditional system ranking.

### 3.5 Robustness Across Test Beds

The average size of the clusters in the gold standard may change from one test bed to another. As this affects the Purity and Inverse Purity trade-off, the same clustering system may obtain a different balance between both metrics in different corpora; and this may produce contradictory evaluation results when comparing systems across different corpora, even for the same $\alpha$ value.

For instance, in the WePS-1b test bed (Artiles et al., 2007), $B_1$ substantially outperforms $B_{100}$ (0.58 vs. 0.49 using $F_{\alpha=0.5}$). In the WePS-2 data set (Artiles et al., 2009), however, $B_{100}$ outperforms $B_1$ (0.53 versus 0.34). The reason is that singletons are less common in WePS-2. In other words, the comparison between $B_{100}$ and $B_1$ depends both on the $\alpha$ value and of the particular distribution of reference cluster sizes in the test bed.

Our point is that system improvements that are robust across $\alpha$ values (which is not the case of $B_1$ and $B_{100}$) should not be affected by this phenomenon. Therefore, estimating the robustness of system improvements to changes in $\alpha$ should prevent reaching contradictory results for different test beds. Indeed, evidence for this is presented in Section 7.

| $F_{\alpha=0.5}$ | | $F_{0.2}$ | |
|---|---|---|---|
| Ranked systems | F result | Ranked systems | F result |
| $S_1$ | 0.78 | $S_1$ | 0.83 |
| $S_2$ | 0.75 | $S_3$ | 0.78 |
| $S_3$ | 0.75 | $S_2$ | 0.77 |
| $S_4$ | 0.67 | $S_6$ | 0.76 |
| $S_5$ | 0.66 | $S_5$ | 0.73 |
| $S_6$ | 0.65 | $S_8$ | 0.73 |
| $S_7$ | 0.62 | $S_{11}$ | 0.71 |
| $\mathbf{B_1}$ | 0.61 | $S_7$ | 0.67 |
| $S_8$ | 0.61 | $S_{14}$ | 0.66 |
| $S_9$ | 0.58 | $S_{15}$ | 0.66 |
| $S_{10}$ | 0.58 | $S_{12}$ | 0.65 |
| $S_{11}$ | 0.57 | $S_9$ | 0.64 |
| $S_{12}$ | 0.53 | $S_{13}$ | 0.63 |
| $S_{13}$ | 0.49 | $S_4$ | 0.62 |
| $S_{14}$ | 0.49 | $S_{10}$ | 0.6 |
| $S_{15}$ | 0.48 | $\mathbf{B_{100}}$ | 0.58 |
| $\mathbf{B_{100}}$ | 0.4 | $S_{16}$ | 0.56 |
| $S_{16}$ | 0.4 | $\mathbf{B_1}$ | 0.49 |

Table 2: WePS-1b system ranking according to $F_{\alpha=0.5}$ vs $F_{\alpha=0.2}$ using Purity and Inverse Purity

## 4. Proposal

Our primary motivation in this article is to quantify the robustness across $\alpha$ values in order to complement the information given by traditional system rankings. To this end we introduce in this section the Unanimous Improvement Ratio.

### 4.1 Unanimous Improvements

The problem of combining evaluation metrics is closely related with the theory of conjoint measurement (see Section 5.1 for a detailed discussion). Van Rijsbergen (1974) argued that it is not possible to determine empirically which metric combining function is the most adequate in the context of Information Retrieval evaluation. However, starting from the measurement theory principles, van Rijsbergen described the set of properties that a metric combining function should satisfy. This set includes the *Independence* axiom (also called *Single Cancellation*), from which the *Monotonicity* property derives. The Monotonicity property states that the quality of a system that surpasses or equals another one according to all metrics is necessarily equal or better than the other. In other words, one system is better than the other with no dependence whatsoever on how the relative importance of each metric is set.

We will define a combination procedure for metrics, *Unanimous Improvement*, which is based on this property:

$$Q_X(a) \geq_\forall Q_X(b) \text{ if and only if } \forall x \in X.Q_x(a) \geq Q_x(b)$$

where $Q_X(a)$ is the quality of $a$ according to a set of metrics $X$.

This relationship has no dependence on how metrics are scaled or weighted, or on their degree of correlation in the metric set. Equality ($=_\forall$) can be derived directly from $\geq_\forall$: The unanimous equality implies that both systems obtain the same score for all metrics:

$$Q_X(a) =_\forall Q_X(b) \equiv (Q_X(a) \geq_\forall Q_X(b)) \wedge (Q_X(b) \geq_\forall Q_X(a))$$

The strict unanimous improvement implies that one system improves the other strictly at least for one metric, and it is not improved according to any metric:

$$Q_X(a) >_\forall Q_X(b) \equiv (Q_X(a) \geq_\forall Q_X(b)) \wedge \neg(Q_X(a) =_\forall Q_X(b)) \equiv$$

$$(Q_X(a) \geq_\forall Q_X(b)) \wedge \neg(Q_X(b) \geq_\forall Q_X(a))$$

Non comparability $\parallel$ can also derived from here: it occurs when some metrics favor one system and some other metrics favor the other. We refer to this cases as *metric-biased improvements*.

$$Q_X(a)\parallel_\forall Q_X(b) \equiv \neg(Q_X(a) \geq_\forall Q_X(b)) \wedge \neg(Q_X(b) \geq_\forall Q_X(a))$$

The theoretical properties of the Unanimous Improvement are described in depth in Section 5.2. The most important property is that the Unanimous Improvement is the only relational structure that does not depend on relative metric weightings, while satisfying the Independence (Monotonicity) axiom. In other words, we can claim that: *A system improvement according to a metric combining function does not depend whatsoever on metric weightings if and only if there is no quality decrease according to any individual metric.* The theoretical justification of this assertion is developed in Section 5.2.1.

## 4.2 Unanimous Improvement Ratio

According to the Unanimous Improvement, our unique observable over each test case is a three-valued function (unanimous improvement, equality or biased improvement). We need, however, a way of quantitatively comparing systems.

Given two systems, $a$ and $b$, and the Unanimous Improvement relationship over a set of test cases $T$, we have samples where $a$ improves $b$ ($Q_X(a) \geq_\forall Q_X(b)$), samples where $b$ improves $a$ ($Q_X(b) \geq_\forall Q_X(a)$) and also samples with biased improvements ($Q_X(a)\parallel_\forall Q_X(b)$). We will refer to these sets as $T_{a \geq_\forall b}$, $T_{b \geq_\forall a}$ and $T_{a\parallel_\forall b}$, respectively. The *Unanimous Improvement Ratio* (UIR) is defined according to three formal restrictions:

| | Precision | | Recall | | | |
|:---:|:---:|:---:|:---:|:---:|:---:|:---:|
| Test cases $\in T$ | System A | System B | System A | System B | $A \geq_\forall B$ | $B \geq_\forall A$ |
| 1 | 0.5 | 0.5 | 0.5 | 0.5 | YES | YES |
| 2 | 0.5 | 0.5 | 0.2 | 0.2 | YES | YES |
| 3 | **0.5** | 0.4 | 0.2 | 0.2 | YES | NO |
| 4 | 0.6 | 0.6 | **0.4** | 0.3 | YES | NO |
| 5 | **0.7** | 0.6 | **0.5** | 0.4 | YES | NO |
| 6 | **0.3** | 0.1 | **0.5** | 0.4 | YES | NO |
| 7 | 0.4 | **0.5** | 0.5 | **0.6** | NO | YES |
| 8 | 0.4 | **0.6** | 0.5 | **0.6** | NO | YES |
| 9 | **0.3** | 0.1 | 0.5 | **0.6** | NO | NO |
| 10 | 0.2 | **0.4** | **0.5** | 0.3 | NO | NO |

Table 3: Example of experiment input to compute UIR

1. $\text{UIR}(a, b)$ should decrease with the number of biased improvements ($T_{a\|_\forall b}$). In the boundary condition where all samples are biased improvements ($T_{a\|_\forall b} = T$), then $\text{UIR}(a, b)$ should be 0.

2. If $a$ improves $b$ as much as $b$ improves $a$ ($T_{a \geq_\forall b} = T_{b \geq_\forall a}$) then $\text{UIR}(a, b) = 0$.

3. Given a fixed number of biased improvements ($T_{a\|_\forall b}$), $\text{UIR}(a, b)$ should be proportional to $T_{a \geq_\forall b}$ and inversely proportional to $T_{b \geq_\forall a}$.

Given these restrictions, we propose the following UIR definition:

$$\text{UIR}_{X,T}(a, b) = \frac{|T_{a \geq_\forall b}| - |T_{b \geq_\forall a}|}{|T|} =$$

$$\frac{|t \in T/Q_X(a) \geq_\forall Q_X(b)| - |t \in T/Q_X(b) \geq_\forall Q_X(a)|}{|T|}$$

which can be alternatively formulated as:

$$\text{UIR}_{X,T}(a, b) = \text{P}(a \geq_\forall b) - \text{P}(b \geq_\forall a)$$

where these probabilities are estimated in a frequentist manner.

UIR range is $[-1, 1]$ and is not symmetric: $\text{UIR}_{X,T}(a, b) = -\text{UIR}_{X,T}(b, a)$. As an illustration of how UIR is computed, consider the experiment outcome in Table 3. Systems A and B are compared in terms of precision and recall for 10 test cases. For test case 5, for instance, A has an unanimous improvement over B: it is better both in terms of precision ($0.7 > 0.6$) and recall ($0.5 > 0.4$). From the table, UIR value is:

$$\text{UIR}_{X,T}(A, B) = \frac{|T_{A \geq_\forall B}| - |T_{B \geq_\forall A}|}{|T|} = \frac{6 - 4}{10} = 0.2 = -\text{UIR}_{X,T}(B, A)$$

UIR has two formal limitations. First, it is not *transitive* (see Section 5.2). Therefore, it is not possible to define a linear system ranking based on UIR. This is, however, not necessary: UIR is not meant to provide a ranking, but to complement the ranking provided by the F-measure (or other metric combining function), indicating how robust results are to changes in $\alpha$. Section 6.4 illustrates how UIR can be integrated with the insights provided by a system ranking.

The second limitation is that UIR does not consider improvement ranges; therefore, it is less sensitive than the F-measure. Our empirical results, however, show that UIR is sensitive enough to discriminate robust improvements versus metric-biased improvements; and in Section 8 we make an empirical comparison of our non-parametric definition of UIR with a parametric version, with results that make the non-parametric definition preferable.

## 5. Theoretical Foundations

In this section we discuss the theoretical foundations of the Unanimous Improvement Ratio in the framework of the Conjoint Measurement Theory. Then we proceed to describe the formal properties of UIR and their implications from the point of view of the evaluation methodology. Readers interested solely in the practical implications of using UIR may proceed directly to Section 6.

### 5.1 Conjoint Measurement Theory

The problem of combining evaluation metrics is closely related with the *Conjoint Measurement Theory*, which was independently discovered by the economist Debreu (1959) and the mathematical psychologist R. Duncan Luce and statistician John Tukey (Luce & Tukey, 1964). The Theory of Measurement defines the necessary conditions to state an homomorphism between an empirical relational structure (e.g. "John is bigger than Bill") and a numeric relational structure ("John's height is 1.79 meters and Bill's height is 1.56 meters"). In the case of the Conjoint Measurement Theory, the relational structure is factored into two (or more) ordered substructures (e.g. "height and weight").

In our context, the numerical structures are given by the evaluation metric scores (e.g. Purity and Inverse Purity). However, we do not have an empirical quality ordering for clustering systems. Different human assessors could assign more relevance to Purity than to Inverse Purity or viceversa. Nevertheless, the Conjoint Measurement Theory does provide mechanisms that state what kind of numerical structures can produce an homomorphism assuming that the empirical structure satisfies certain axioms. Van Rijsbergen (1974) used this idea to analyze the problem of combining evaluation metrics. These axioms shape an *additive conjoint structure*. Being $(R, P)$ the quality of a system according to two evaluation metrics $R$ and $P$, these axioms are:

**Connectedness:** All systems should be comparable to each other. Formally: $(R, P) \geq (R', P')$ or $(R', P') \geq (R, P)$.

**Transitivity:** $(R, P) \geq (R', P')$ and $(R', P') \geq (R'', P'')$ implies that $(R, P) \geq (R'', P'')$. The axioms Transitivity and Connectedness shape a *weak order*.

**Thomsen condition:** $(R_1, P_3) \sim (R_3, P_2)$ and $(R_3, P_1) \sim (R_2, P_3)$ imply that $(R_1, P_1) \sim (R_2, P_2)$ (where $\sim$ indicates equal effectiveness).

**Independence:** *"The two components contribute their effects independently to the effectiveness"*. Formally, $(R_1, P) \geq (R_2, P)$ implies that $(R_1, P') \geq (R_2, P')$ for all $P'$, and $(R, P_1) \geq (R, P_2)$ implies that $(R', P_2) \geq (R', P_2)$ for all $R'$. This property implies *Monotonicity* (Narens & Luce, 1986) which states that an improvement in both metrics necessarily produces an improvement according to the metric combining function.

**Restricted Solvability:** A property which is *"... concerned with the continuity of each component. It makes precise what intuitively we would expect when considering the existence of intermediate levels"*. Formally: whenever $(R_1, P') \geq (R, P) \geq (R_2, P')$ then exists R' such that $(R', P') = (R, P)$.

**Essential Components:** *"Variation in one while leaving the other constant gives a variation in effectiveness"*. There exists $R$, $R'$ and $P$ such that it is not the case that $(R, P) = (R', P)$; and there exists $P$, $P'$ and $R$ such that it is not the case that $(R, P) = (R, P')$.

**Archimedean Property:** which *"merely ensures that the intervals on a component are comparable"*.

The F-measure proposed by van Rijsbergen (1974) and the arithmetic mean of P,R satisfy all these axioms. According to these restrictions, indeed, an unlimited set of acceptable combining functions for evaluation metrics can be defined. The F relational structure, however, satisfies another property which is not satisfied by other functions such as the arithmetic mean. This property is the *Decreasing Marginal Effectiveness*. The basic idea is that increasing one unit in one metric and decreasing one unit in the other metric can improve the overall quality (i.e. if the first metric has more weight in the combining function), but this does not imply that a great loss in one metric can be compensated by a great increase in the other. It can be defined as:

$$\forall R, P > 0, \exists n > 0 \text{ such that } ((P + n, R - n) < (R, P))$$

According to this, high values in both metrics are required to obtain a high overall improvement. This makes measures observing this property - such as F - more robust to arbitrary metric weightings.

## 5.2 Formal Properties of the Unanimous Improvement

The Unanimous Improvement $\geq_{\forall x}$ trivially satisfies most of the desirable properties proposed by van Rijsbergen (1974) for metric combining functions: transitivity, independence, Thomsen condition, Restricted Solvability, Essential Components and Decreasing Marginal Effectiveness; the exception being the connectedness property[4]. Given that the non comparability $\|_{\forall}$ (biased improvements, see Section 4.1) is derived from the Unanimous Improvement, it is possible to find system pairs where neither $Q_X(a) \geq_{\forall} Q_X(b)$ nor $Q_X(b) \geq_{\forall} Q_X(a)$ hold. Therefore, *Connectedness* is not satisfied.

Formally, the limitation of the Unanimous Improvement is that it does not represent a *weak order*, because it cannot satisfy *Transitivity* and *Connectedness* simultaneously. Let us elaborate on this issue.

---

4. For the sake of simplicity, we consider here the combination of two metrics $(R, P)$.

| Systems | Metric $x_1$ | Metric $x_2$ |
|---------|--------------|--------------|
| A | 0.5 | 0.5 |
| B | 0.6 | 0.4 |
| C | 0.45 | 0.45 |

Table 4: A counter sample for *Transitivity* in Unanimous Improvement

We could satisfy *Connectedness* by considering that biased improvements represent equivalent system pairs ($=_\forall$). But in this case, *Transitivity* would not be satisfied. See, for instance, Table 4. According to the table:

$$Q_X(B)\|_\forall Q_X(A) \text{ and } Q_X(C)\|_\forall Q_X(B)$$

Therefore, considering that $\|_\forall$ represents equivalence, we have:

$$Q_X(B) \geq_\forall Q_X(A) \text{ and } Q_X(C) \geq_\forall Q_X(B)$$

but not

$$Q_X(C) \geq_\forall Q_X(A)$$

In summary, we can choose to satisfy transitivity or connectedness, but not both: the Unanimous Improvement can not derive a *weak order*.

### 5.2.1 Uniqueness of the Unanimous Improvement

The Unanimous Improvement has the interesting property that is does not contradict any evaluation result given by the F-measure, regardless of the $\alpha$ value used in F:

$$Q_X(a) \geq_\forall Q_X(b) \rightarrow \forall \alpha F_\alpha(a) \geq F_\alpha(b)$$

This is due to the fact that the F-measure (for any $\alpha$ value) satisfies the *monotonicity* axiom, in which the Unanimous Improvement is grounded. This property is essential for the purpose of checking the robustness of system improvements across $\alpha$ values. And crucially, the Unanimous Improvement is *the only function* that satisfies this property. More precisely, the Unanimous Improvement is the only relational structure that, while satisfying monotonicity, does not contradict any Additive Conjoint Structure (see Section 5.1).

In order to prove this assertion, we need to define the concept of *compatibility* with any additive conjoint structure. Let $\geq_{add}$ be any additive conjoint structure and let $\geq_R$ be any relational structure. We will say that $\geq_R$ is *compatible* with any conjoint structure if and only if:

$$\forall \langle a, b, \geq_{add} \rangle . (Q_X(a) \geq_R Q_X(b)) \rightarrow (Q_X(a) \geq_{add} Q_X(b))$$

In other words: if $\geq_R$ holds, then any other additive conjoint holds. We want to prove that the unanimous improvement is the only relation that satisfies this property; therefore, we have to prove that if $\geq_R$ is a monotonic and compatible relational structure, then it necessarily matches the unanimous improvement definition:

$\geq_R$ is monotonic and compatible $\implies (Q_X(a) \geq_R Q_X(b) \leftrightarrow x_i(a) \geq x_i(b) \forall x_i \in X)$

which can be split in:

(1) $\geq_R$ monotonic and compatible $\implies (Q_X(a) \geq_R Q_X(b) \leftarrow x_i(a) \geq x_i(b) \forall x_i \in X)$

(2) $\geq_R$ monotonic and compatible $\implies (Q_X(a) \geq_R Q_X(b) \rightarrow x_i(a) \geq x_i(b) \forall x_i \in X)$

Proving (1) is immediate, since the rightmost component corresponds with the monotonicity property definition. Let us prove (2) by *reductio ad absurdum*, assuming that there exists a relational structure $\geq_o$ such that:

$$(\geq_o \text{ monotonic and compatible}) \wedge (Q_X(a) \geq_o Q_X(b)) \wedge (\exists x_i \in X. x_i(a) < x_i(b))$$

In this case, we could define an *additive conjoint structure* over the combined measure $Q'_X(a) = \alpha_1 x_1(a) + .. \alpha_i x_i(a) .. + \alpha_n x_n(a)$ with $\alpha_i$ big enough such that $Q'_X(a) < Q'_X(b)$. The $Q'$ *additive conjoint structure* would contradict $\geq_o$. Therefore, $\geq_o$ would not be compatible (contradiction). In conclusion, predicate (2) is true and the Unanimous Improvement $\geq_{\forall X}$ is the only monotonic and compatible relational structure.

An interesting corollary can be derived from this analysis. If the Unanimous Improvement is the only *compatible* relational structure, then we can formally conclude that the measurement of system improvements without dependence on metric weighting schemes can not derive a *weak order* (i.e. one that satisfies both transitivity and connectedness). This corollary has practical implications: *it is not possible to establish a system ranking which is independent on metric weighting schemes.*

A natural way to proceed is, therefore, to use the unanimous improvement as an addition to the standard F-measure (for a suitable $\alpha$ value) which provides additional information about the robustness of system improvements across $\alpha$ values.

## 6. F versus UIR: Empirical Study

In this Section we perform a number of empirical studies on the WePS corpora in order to find out how UIR behaves in practice. First, we focus on a number of empirical results that show how UIR rewards robustness across $\alpha$ values, and how this information is complementary to the information provided by F. Second, we examine to what extent – and why – F and UIR are correlated.

### 6.1 UIR: Rewarding Robustness

Figure 4 shows three examples of system comparisons in WePS-1b corpus using the metrics Purity and Inverse Purity. Each curve represents the $F_\alpha$ value obtained for one system according to different $\alpha$ values. System S6 (black curves) is compared with S10, S9 and S11 (grey curves) in each of the three graphs. In all cases there is a similar quality increase according to $F_{\alpha=0.5}$; UIR, however, ranges between 0.32 and 0.42, depending on how robust the difference is to changes in $\alpha$. The highest difference in UIR is for the (S6,S11) system pair (rightmost graph), because these systems do not swap their $F_\alpha$ values for any $\alpha$ value.

|  | Improvements for all $\alpha$ (28 system pairs) | Other cases (125 system pairs) |
|---|---|---|
| $\lvert \triangle F_{\alpha=0.5} \rvert$ | 0.12 | 0.13 |
| $\lvert \text{UIR} \rvert$ | 0.53 | 0.14 |

Table 5: UIR and $F_{\alpha=0.5}$ increase when F increases for all $\alpha$ values

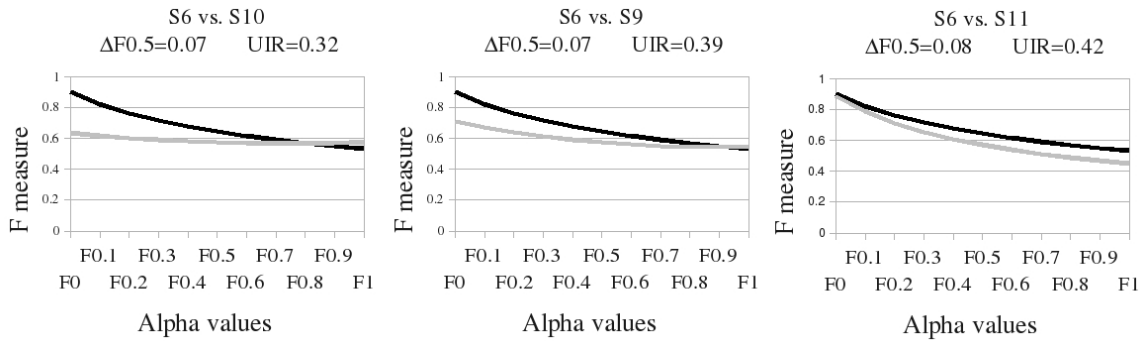

Figure 4: F-measure vs. UIR: rewarding robustness

The smallest UIR value is for (S6,S10), where S6 is better than S10 for $\alpha$ values below 0.8, and worse when $\alpha$ is larger. This comparison illustrates how UIR captures, for similar increments in F, which ones are less dependent of the relative weighting scheme between precision and recall.

Let us now consider all two-system combinations in the WePS-1b corpus, dividing them in two sets: (i) system pairs for which $F_\alpha$ increases for all $\alpha$ values (i.e. both Purity and Inverse Purity increases), and (ii) pairs for which the relative system's performance swaps at some $\alpha$ value; i.e. $F_\alpha$ increases for some $\alpha$ values and decreases for the rest.

One would expect that the average increase in $F_\alpha$ should be larger for those system pairs where one beats the other for every $\alpha$ value. Surprisingly, this is not true: Table 5 shows the average increments for UIR and $F_{\alpha=0.5}$ for both sets. UIR behaves as expected: its average value is substantially larger for the set where different $\alpha$ do not lead to contradictory results (0.53 vs. 0.14). But the average relative increase of $F_{\alpha=0.5}$, however, is very similar in both sets (0.12 vs. 0.13).

The conclusion is that a certain $F_{\alpha=0.5}$ improvement range does not say anything about whether both Purity and Inverse Purity are being simultaneously improved or not. In other words: no matter how large is a measured improvement in $F$ is, it can still be extremely dependent on how we are weighting the individual metrics in that measurement.

This conclusion can be corroborated by considering independently both metrics (Purity and Inverse Purity). According to the statistical significance of the improvements for independent metrics, we can distinguish three cases:

1. *Opposite significant improvements*: One of the metrics (Purity or Inverse Purity) increases and the other decreases, and both changes are statistically significant.

| | Significant concordant improvements 53 pairs | Significant opposite improvements 89 pairs | Non significant improvements 11 pairs |
|---|---|---|---|
| $\mid \triangle F_{\alpha=0.5} \mid$ | 0.11 | 0.15 | 0.05 |
| $\mid$UIR$\mid$ | 0.42 | 0.08 | 0.027 |

Table 6: UIR and $F_{\alpha=0.5}$ increases vs. statistical significance tests

2. *Concordant significant improvements*: Both metrics improve significantly or at least one improves significantly and the other does not decrease significantly.

3. *Non-significant improvements*: There is no statistically significant differences between both systems for any metric.

We use the Wilcoxon test with $p < 0.05$ to detect statistical significance. Table 6 shows the average UIR and $\mid \triangle F_{\alpha=0.5} \mid$ values in each of the three cases. Remarkably, the $F_{\alpha=0.5}$ average increase is even larger for the opposite improvements set (0.15) than for the concordant improvements set (0.11). According to these results, it would seem that $F_{\alpha=0.5}$ rewards individual metric improvements which are obtained at the cost of (smaller) decreases in the other metric. UIR, on the other hand, has a sharply different behavior, strongly rewarding the concordant improvements set (0.42 versus 0.08).

All these results confirm that UIR provides essential information about the experimental outcome of two-system comparisons, which is not provided by the main evaluation metric $F_\alpha$.

## 6.2 Correlation Between F and UIR

The fact that UIR and F offer different information about the outcome of an experiment does not imply that UIR and F are orthogonal; in fact, there is some correlation between both values.

Figure 5 represents $F_{\alpha=0.5}$ differences and UIR values for each possible system pair in the WePS-1 test bed. The general trends are (i) high UIR values imply a positive difference in F (ii) high $\mid \triangle F_{0.5} \mid$ values do not imply anything on UIR values; (iii) low UIR do not seem to imply anything on $\mid \triangle F_{0.5} \mid$ values. Overall, the figure suggest a triangle relationship, which gives a Pearson correlation of 0.58.

### 6.2.1 REFLECTING IMPROVEMENT RANGES

When there is a consistent difference between two systems for most $\alpha$ values, UIR rewards larger improvement ranges. Let us illustrate this behavior considering three sample system pairs taken from the WePS-1 test bed.

Figure 6 represents the $F_{\alpha \in [0,1]}$ values for three system pairs. In all cases, one system improves the other for all $\alpha$ values. However, UIR assigns higher values to larger improvements in F (larger distance between the black and the grey curves). The reason is that a larger average improvement over test cases makes less likely the cases where individual test cases (which are the ones that UIR considers) contradict the average result.

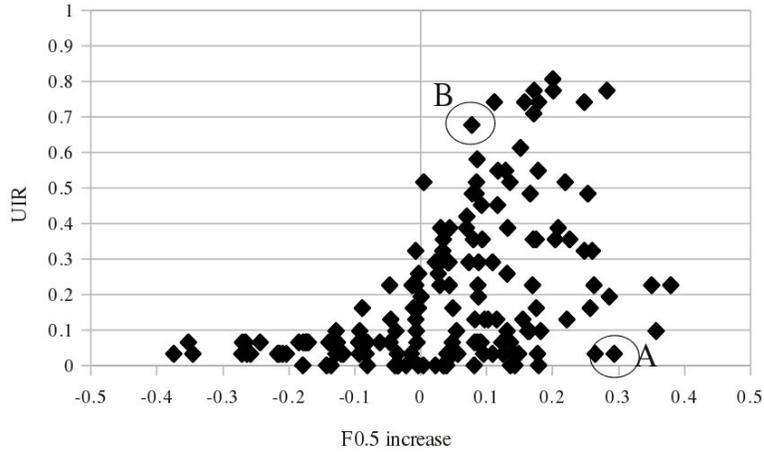

Figure 5: $|\triangle F_{0,.5}|$ vs UIR

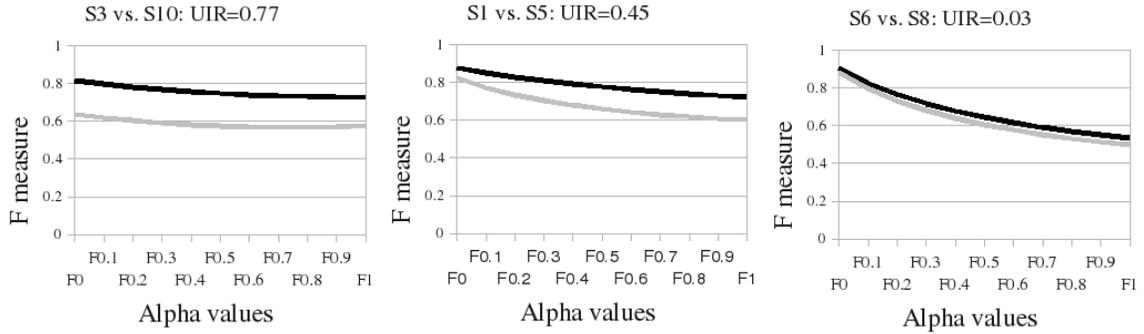

Figure 6: F vs. UIR: reflecting improvement ranges

Another interesting finding is that, when both metrics are improved, the metric that has the weakest improvement determines the behavior of UIR. Figure 7 illustrates this relationship for the ten system pairs with a largest improvement; the Pearson correlation in this graph is 0.94. In other words, when both individual metrics improve, UIR is sensitive to the weakest improvement.

### 6.2.2 Analysis of boundary cases

In order to have a better understanding of the relationship between UIR and F, we will now examine in detail two cases of system improvements in which UIR and F produce drastically different results. These two cases are marked as A and B in Figure 5.

The point marked as case A in the Figure corresponds with the comparison of systems $S_1$ and $S_{15}$. There exists a substantial (and statistically significant) difference between both systems according to $F_{\alpha=0.5}$. However, UIR has a low value, i.e., the improvement is not robust to changes in $\alpha$ according to UIR.

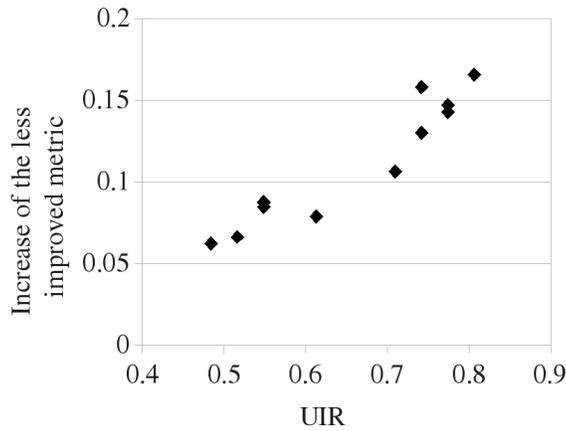

Figure 7: Correlation between UIR and the weakest single metric improvement.

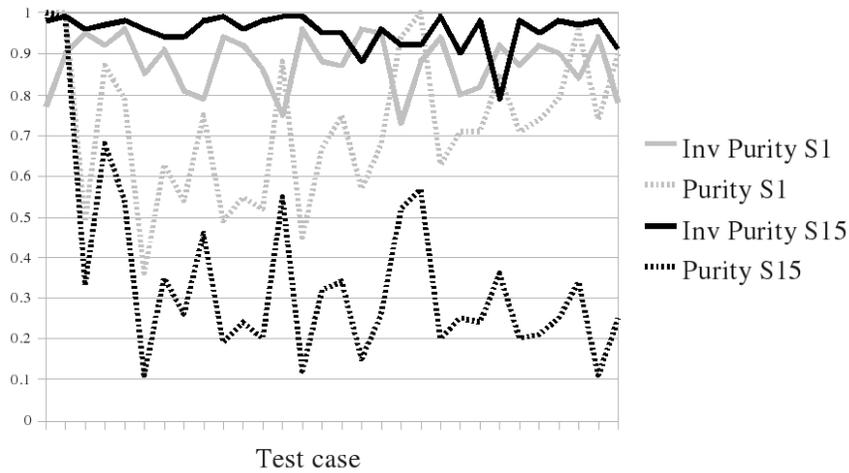

Figure 8: Purity and Inverse Purity per test case, systems $S_1$ and $S_1 5$

A visual explanation of these results can be seen in Figure 8. It shows the Purity and Inverse Purity results of systems $S_1, S_{15}$ for every test case. In most test cases, $S_1$ has an important advantage in Purity at the cost of a slight – but consistent – loss in Inverse Purity. Given that $F_{\alpha=0.5}$ compares Purity and Inverse Purity ranges, it states that there exists an important and statistically significant improvement from $S_{15}$ to $S_1$. However, the slight but consistent decrease in Inverse Purity affects UIR, which decreases because in most test cases the improvements in F are metric biased ($\|_\forall$ in our notation).

Case B (see Figure 9) is the opposite example: there is a small difference between systems S8 and S12 according to $F_{\alpha=0.5}$, because differences in both Purity and Inverse Purity are also small. S8, however, gives small but consistent improvements both for Purity and Inverse Purity (all test cases to the right of the vertical line in the figure); these are unanimous improvements. Therefore, UIR considers that there exists a robust overall improvement in this case.

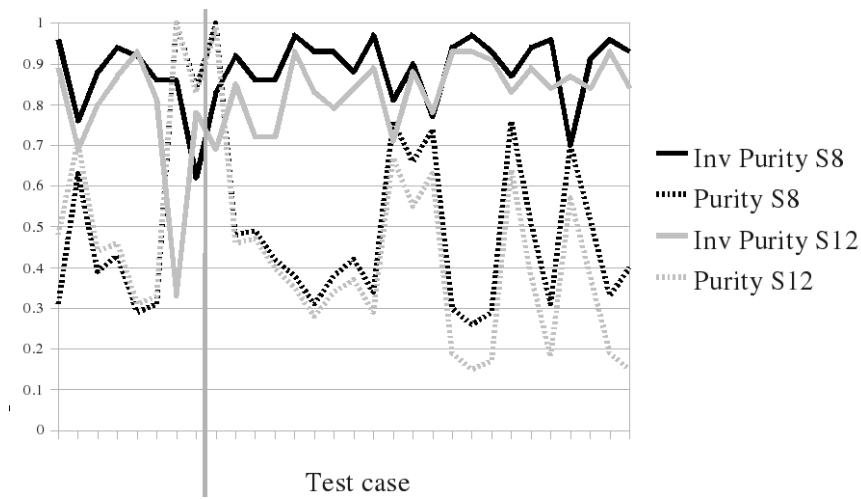

Figure 9: Purity and Inverse Purity per test case, systems $S_12$ and $S_8$

Again, both cases show how UIR gives additional valuable information on the comparative behavior of systems.

## 6.3 A Significance Threshold for UIR

We mentioned earlier that UIR has a parallelism with statistical significance tests, which are typically used in Information Retrieval to estimate the probability $p$ that an observed difference between two systems is obtained by chance, i.e., the difference is an artifact of the test collection rather than a true difference between the systems. When computing statistical significance, it is useful to establish a threshold that allows for a binary decision; for instance, a result is often said to be statistically significant if $p < 0.05$, and not significant otherwise. Choosing the level of significance is arbitrary, but it nevertheless helps reporting and summarizing significance tests. Stricter thresholds increase confidence of the test, but run an increased risk of failing to detect a significant result.

The same situation applies to UIR: we would like to establish an UIR threshold that decides whether an observed difference is reasonably robust to changes in $\alpha$. How can we set such a threshold? We could be very restrictive and decide, for instance, that an improvement is significantly robust when UIR $\geq 0.75$. This condition, however, is so hard that it would never be satisfied in practice, and therefore the UIR test would not be informative. On the other hand, if we set a very permissive threshold it will be satisfied by most system pairs and, again, it will not be informative. The question now is whether there exists a threshold for UIR values such that obtaining a UIR above the threshold guarantees that an improvement is robust, and, at the same time, is not too strong to be satisfied in practice.

Given the set of two-system combinations for which UIR surpasses a certain candidate threshold, we can think of some desirable features:

1. It must be able to differentiate between two types of improvements (*robust* vs. *non-robust*); in other words, if one of the two types is usually empty or almost empty, the threshold is not informative.

2. The *robust* set should contain a high ratio of two-system combinations such that the average $F_\alpha$ increases for all $\alpha$ values ($F_\alpha(a) > F_\alpha(b) \forall \alpha$).

3. The *robust* set should contain a high ratio of *significant concordant improvements* and a low ratio of *significant opposite improvements* (see Section 6.1).

4. The *robust* set should contain a low ratio of cases where F contradicts UIR (the dots in Figure 5 in the region $|\triangle F_{0,.5}| < 0$).

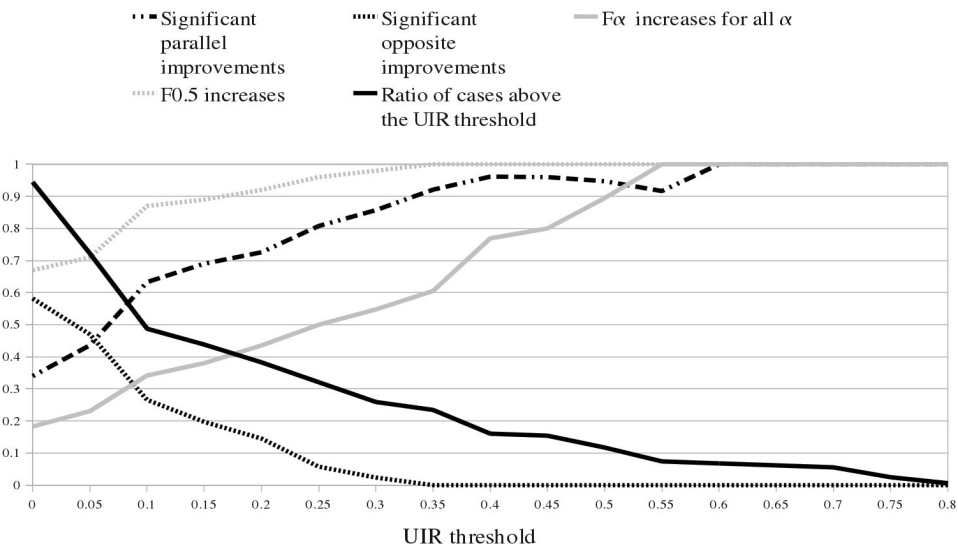

Figure 10: Improvement detected across UIR thresholds

Figure 10 shows how these conditions are met for every threshold in the range $[0, 0.8]$. A UIR threshold of 0.25 accepts around 30% of all system pairs, with a low (4%) ratio of *significant opposite improvements* and a high (80%) ratio of *significant concordant improvements*. At this threshold, in half of the *robust* cases $F_\alpha$ increases for all $\alpha$ values, and in most cases (94%) $F_{\alpha=0.5}$ increases. It seems, therefore, that UIR $\geq 0.25$ can be a reasonable threshold, at least for this clustering task. Note, however, that this is a rough rule of thumb that should be revised/adjusted when dealing with clustering tasks other than WePS.

## 6.4 UIR and System Rankings

All results presented so far are focused on pairwise system comparisons, according to the nature of UIR. We now turn to the question of how can we use UIR as a component in the analysis of the results of an evaluation campaign.

In order to answer this question we have applied UIR to the results of the WePS-2 evaluation campaign (Artiles et al., 2009). In this campaign, the best runs for each system were ranked according to Bcubed precision and recall metrics, combined with $F_{\alpha=0.5}$. In addition to all participant systems, three baseline approaches were included in the ranking:

all documents in one cluster ($B_{100}$), each document in one cluster ($B_1$) and the union of both ($B_{COMB}$)[5].

Table 7 shows the results of applying UIR to the WePS-2 participant systems. $\alpha$-robust improvements are represented in the third column ("improved systems"): for every system, it displays the set of systems that it improves with UIR $\geq 0.25$. The fourth column is the *reference system*, which is defined as follows: given a system $a$, its reference system is the one that improves $a$ with maximal UIR:

$$S_{ref}(a) = Argmax_S(\text{UIR}(S, a))$$

In other words, $S_{ref}(a)$ represents the system with which $a$ should be replaced in order to robustly improve results across different $\alpha$ values. Finally, the last column ("UIR for the reference system") displays the UIR between the system and its reference ($\text{UIR}(S_{ref}, S_i)$).

Note that UIR adds new insights into the evaluation process. Let us highlight two interesting facts:

- Although the three top-scoring systems (S1, S2, S3) have a similar performance in terms of F (0.82, 0.81 and 0.81), $S_1$ is consistently the best system according to UIR, because it is the reference for 10 other systems (S2, S4, S6, S8, S12, S13, S14, S15, S16 and the baseline $B_1$). In contrast, S2 is reference for S7 only, and S3 is reference for S11 only. Therefore, F and UIR together strongly point towards S1 as the best system, while F alone was only able to discern a set of three top-scoring systems.

- Although the non-informative baseline $B_{100}$ (all documents in one cluster) is better than five systems according to F, this improvement is not robust according to UIR. Note that UIR will signal near-baseline behaviors in participant systems with a low value, while they can receive a large F depending on the nature of the test collection: when the average cluster is large or small, systems that tend to cluster everything or nothing can be artificially rewarded. This is, in our opinion, a substantial improvement over using F alone.

## 7. UIR as Predictor of the Stability of Results across Test Collections

A common issue when evaluating systems that deal with Natural Language is that results on different test collections are often contradictory. In the particular case of Text Clustering, a factor that contributes to this problem is that the average size of clusters can vary across different test beds, and this variability modifies the optimal balance between precision and recall. A system which tends to favor precision, creating small clusters, may have good results in a dataset with a small average cluster size and worse results in a test collection with a larger average cluster size.

Therefore, if we only apply $F$ to combine single metrics, we can reach contradictory results over different test beds. As UIR does not depend on metric weighting criteria, our hypothesis is that a high UIR value ensures robustness of evaluation results across test beds.

---

5. See the work of Artiles et al. (2009) for an extended explanation.

| System | $F_{0.5}$ | Improved systems (UIR > 0.25) | Reference system | UIR for the reference system |
|---|---|---|---|---|
| S1 | 0,82 | S2 S4 S6 S7 S8 S11..S17 B$_1$ | - | - |
| S2 | 0,81 | S4 S6 S7 S8 S11..S17 B$_1$ | S1 | 0,26 |
| S3 | 0,81 | S2 S4 S7 S8 S11..S17 B$_1$ | - | - |
| S4 | 0,72 | S11 S13..S17 | S1 | 0,58 |
| S5 | 0,71 | S12..S16 | - | - |
| S6 | 0,71 | S4 S7 S11 S13..S17 B$_1$ | S1 | 0,35 |
| S7 | 0,70 | S11 S13..S17 | S2 | 0,65 |
| S8 | 0,70 | S11..S17 | S1 | 0,74 |
| S9 | 0,63 | S4 S12 S14 S16 | - | - |
| S10 | 0,63 | S12..S16 | - | - |
| S11 | 0,57 | S14..S17 | S3 | 0,68 |
| S12 | 0,53 | S14 S16 | S1 | 0,71 |
| B$_{100}$ | 0,53 | B$_{COMB}$ | - | - |
| S13 | 0,52 | S15 S16 | S1 | 0,9 |
| B$_{COMB}$ | 0,52 | - | B$_{100}$ | 0,65 |
| S14 | 0,42 | - | S1 | 0,9 |
| S15 | 0,41 | S16 | S1 | 0,97 |
| S16 | 0,39 | - | S1 | 1,00 |
| B$_1$ | 0,34 | S17 | S1 | 0,29 |
| S17 | 0,33 | - | S6 | 0,84 |

Table 7: WePS-2 results with Bcubed precision and recall, F and UIR measures

In other words: *given a particular test bed, a high UIR value should be a good predictor that an observed difference between two systems will still hold in other test beds.*

The following experiment is designed to verify our hypothesis. We have implemented four different systems for the WePS problem, all based on an agglomerative clustering algorithm (HAC) which was used by the best systems in WePS-2. Each system employs a certain cluster linkage technique (complete link or single link) and a certain feature extraction criterion (word bigrams or unigrams). For each system we have experimented with 20 stopping criteria. Therefore, we have used 20x4 system variants overall. We have evaluated these systems over WePS-1a, WePS-1b and WePS-2 corpora[6].

The first observation is that, given all system pairs, $F_{\alpha=0.5}$ only gives consistent results for all three test beds in 18% of the cases. For all other system pairs, the best system is different depending of the test collection. A robust evaluation criterion should predict, given a single test collection, whether results will still hold in other collections.

We now consider two alternative ways of predicting that an observed difference (system A is better than system B) in one test-bed will still hold in all three test beds:

- The first is using $F(A) - F(B)$: the larger this value is on the reference test bed, the more likely that $F(A) - F(B)$ will still be positive in a different test collection.

---

6. WEPS-1a was originally used for training in the first WePS campaign, and WePS-1b was used for testing.

- The second is using $UIR(A, B)$ instead of F: the larger UIR is, the more likely that $F(A) - F(B)$ is also positive in a different test bed.

In summary, we want to compare F and UIR as predictors of how robust is a result to a change of test collection. This is how we tested it:

1. We select a reference corpus out of WePS-1a, WePS-1b and WePS-2 test beds.

$$C_{ref} \in \{\text{WePS-1a,WePS-1b,WePS-2}\}$$

2. For each system pair in the reference corpus, we compute the improvement of one system with respect to the other according to F and UIR. We take those system pairs such that one improves the other over a certain threshold $t$. Being $\text{UIR}_C(s_1, s_2)$ the UIR results for systems $s_1$ and $s_2$ in the test-bed C, and being $\text{F}_C(s)$ the results of F for the system $s$ in the test-bed C:

$$S_{UIR,t}(C) = \{(s_1, s_2)|\text{UIR}_C(s_1, s_2) > t\}$$

$$S_{F,t}(C) = \{s_1, s_2|(\text{F}_C(s_1) - \text{F}_C(s_2)) > t)\}$$

For every threshold $t$, $S_{UIR,t}$ and $S_{F,t}$ represent the set of robust improvements as predicted by UIR and F, respectively.

3. Then, we consider the system pairs such that one improves the other according to F for all the three test collections simultaneously.

$$T = \{s_1, s_2|\text{F}_C(s_1) > \text{F}_C(s_2) \forall C\}$$

$T$ is the gold standard to be compared with predictions $S_{UIR,t}$ and $S_{F,t}$.

4. For every threshold $t$, we can compute precision and recall of UIR and F predictions ($S_{UIR,t}(C)$ and $S_{F,t}(C)$) versus the actual set of robust results across all collections ($T$).

$$Precision(S_{UIR,t}(C)) = \frac{|S_{UIR,t}(C) \cap T|}{|S_{UIR,t}|} \quad Recall(S_{UIR,t}(C)) = \frac{|S_{UIR,t}(C) \cap T|}{|T|}$$

$$Precision(S_{F,t}(C)) = \frac{|S_{F,t}(C) \cap T|}{|S_{F,t}(C)|} \quad Recall(S_{F,t}(C)) = \frac{|S_{F,t}(C) \cap T|}{|T|}$$

We can now trace the precision/recall curve for each of the predictors F, UIR and compare their results. Figures 11, 12 and 13, show precision/recall values for F (triangles) and UIR (rhombi); each figure displays results for one of the reference test-beds: WEPS-1a,WEPS-1b and WePS-2[7].

Altogether, the figures show how UIR is much more effective than F as a predictor. Note that F suffers a sudden drop in performance for low recall levels, which suggests that big improvements in F tend to be due to the peculiarities of the test collection rather than to a real superiority of one system versus the other.

---

7. The curve "parametric UIR" refers to an alternative definition of UIR which is explained in Section 8

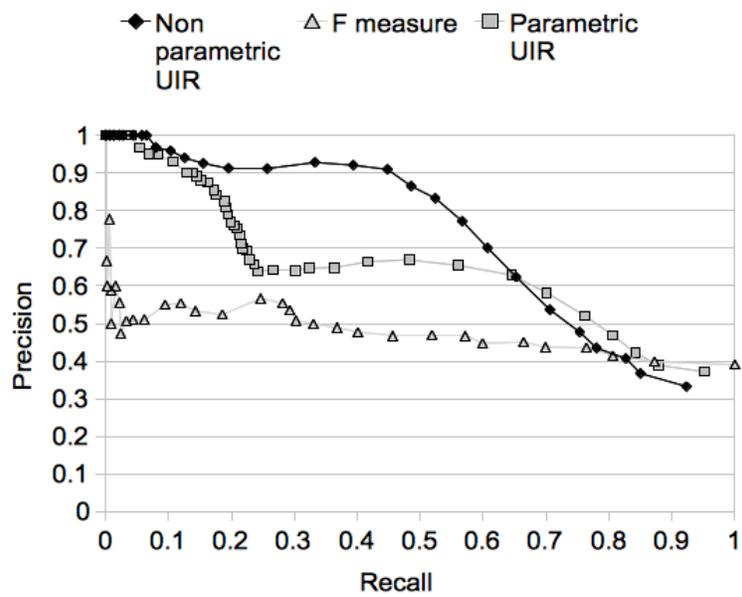

Figure 11: Predictive power of UIR and F from WePS-1a

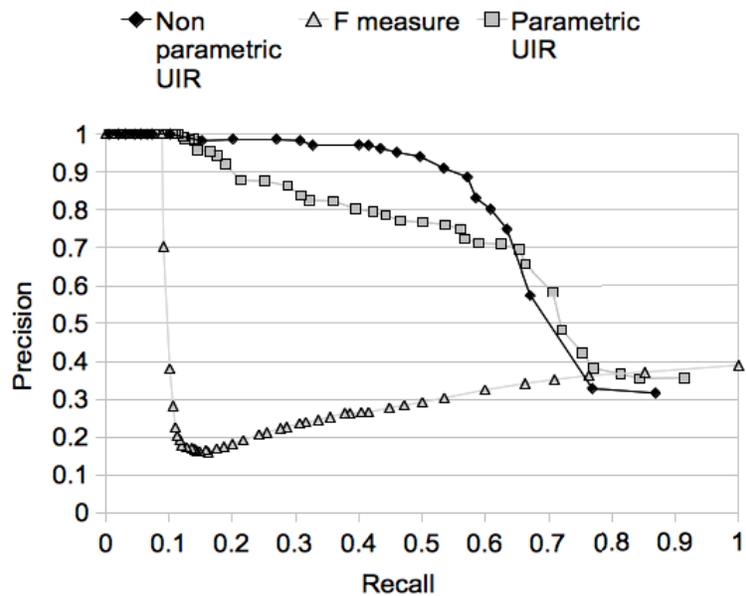

Figure 12: Predictive power of UIR and F from WePS-1b

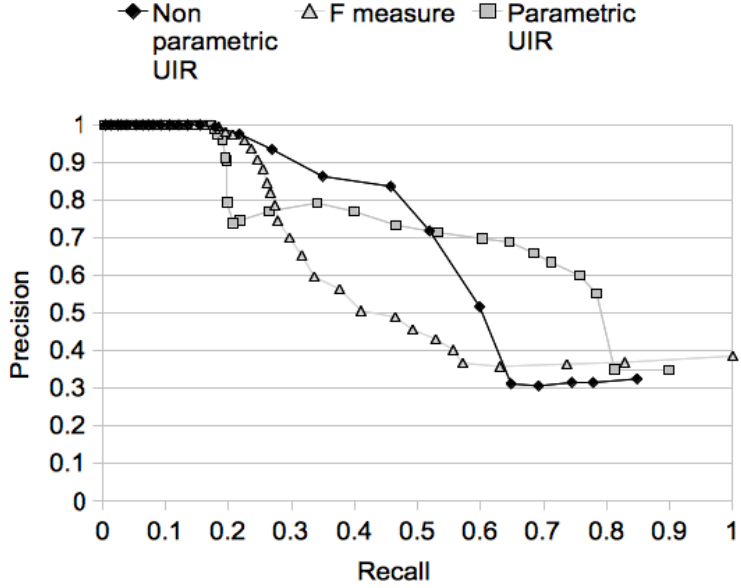

Figure 13: Predictive power of UIR and F from WePS-2

This is, in our opinion, a remarkable result: differences in UIR are better indicators of the reliability of a measured difference in F than the amount of the measured difference. Therefore, UIR is not only useful to know how stable are results to changes in $\alpha$, but also to changes in the test collection, i.e., it is an indicator of how reliable a perceived difference is.

Note that we have not explicitly tested the dependency (and reliability) of UIR results with the number of test cases in the reference collection. However, as working with a collection of less than 30 test cases is unlikely, in practical terms the usability of UIR is granted for most test collections, at least with respect of the number of test cases.

## 8. Parametric versus Non-Parametric UIR

According to our analysis (see Section 5.2), given two measures P and R, the only relational structure over pairs $\langle P_i, R_i \rangle$ that does not depend on weighting criteria is the unanimous improvement:

$$a \geq_\forall b \equiv P_a \geq P_b \wedge R_a \geq R_b$$

When comparing systems, our UIR measure counts the unanimous improvement results across test cases:

$$\text{UIR}_{X,T}(a,b) = \frac{|T_{a \geq_\forall b}| - |T_{b \geq_\forall a}|}{|T|}$$

Alternatively, this formulation can be expressed in terms of probabilities:

$$\text{UIR}_{X,T}(a,b) = \text{Prob}(a \geq_\forall b) - \text{Prob}(b \geq_\forall a)$$

where these probabilities are estimated in a frequentist manner.

As we said, the main drawback of the unanimous improvement is that it is a three-valued function which does not consider metric ranges; UIR inherits this drawback. As a consequence, UIR is less sensitive than other combining schemes such as the F measure. In order to solve this drawback, we could estimate UIR parametrically. However, the results in this section seem to indicate that this is not the best option.

One way of estimating $Prob(a \geq_\forall b)$ and $Prob(b \geq_\forall a)$ consists of assuming that the metric differences $(\Delta P, \Delta R)$ between two systems across test cases follow a normal bivariate distribution. We can then estimate this distribution from the case samples provided in each test bed. After estimating the density function $Prob(\Delta P, \Delta R)$, we can estimate $Prob(a \geq_\forall b)$ as[8]:

$$Prob(a \geq_\forall b) = Prob(\Delta P \geq 0 \wedge \Delta R \geq 0) = \int_{\Delta P = 0, \Delta R = 0}^{\Delta P = 1, \Delta R = 1} Prob(\Delta P, \Delta R) \, d\Delta P \, d\Delta R$$

This expression can be used to compute $\text{UIR}_{X,T}(a,b) = \text{Prob}(a \geq_\forall b) - \text{Prob}(b \geq_\forall a)$, and leads to a parametric version of UIR.

In order to compare the effectiveness of the parametric UIR versus the original UIR, we repeated the experiment described in Section 7, adding UIR$_{\text{param}}$ to the precision/recall curves in Figures 11, 12 and 13. The squares in these figures represent the results for the parametric version of UIR. Note that its behavior lies somewhere between F and the non-parametric UIR: for low levels of recall, it behaves like the original UIR; for intermediate levels, it is in general worse than the original definition but better than F; and in the recall high-end, it overlaps with the results of F. This is probably due to the fact that the parametric UIR estimation considers ranges, and becomes sensitive to the unreliability of high improvements in F.

## 9. Conclusions

Our work has addressed the practical problem of the strong dependency (and usually some degree of arbitrariness) on the relative weights assigned to metrics when applying metric combination criteria, such as $F$.

Based on the theory of measurement, we have established some relevant theoretical results: the most fundamental is that there is only one monotonic relational structure that does not contradict any *Additive Conjoint Structure*, and that this unique relationship is not transitive. This implies that it is not possible to establish a ranking (a complete ordering) of systems without assuming some arbitrary relative metric weighting. A transitive relationship, however, is not necessary to ensure the robustness of specific pairwise system comparisons.

Based on this theoretical analysis, we have introduced the *Unanimous Improvement Ratio* (UIR), which estimates the robustness of measured system improvements across potential metric combining schemes. UIR is a measure complementary to any metric combination scheme and it works similarly to a statistical relevance test, indicating if a perceived difference between two systems is reliable or biased by the particular weighting scheme used to evaluate the overall performance of systems.

---

8. For this computation we have employed the Matlab tool

Our empirical results on the text clustering task, which is particularly sensitive to this problem, confirm that UIR is indeed useful as an analysis tool for pairwise system comparisons: (i) For similar increments in F, UIR captures which ones are less dependent of the relative weighting scheme between precision and recall; (ii) unlike F, UIR rewards system improvements that are corroborated by statistical significance tests over each single measure; (iii) in practice, a high UIR tends to imply a large F increase, while a large increase in F does not imply a high UIR; in other words, a large increase in F can be completely biased by the weighting scheme, and therefore UIR is an essential information to add to F.

When looking at results of an evaluation campaign, UIR has proved useful to (i) discern which is the best system among a set of systems with similar performance according to $F$; (ii) penalize trivial baseline strategies and systems with a baseline-like behavior.

Perhaps the most relevant result is a side effect on how our proposed measure is defined: UIR is a good estimator of how robust a result is to changes in the test collection. In other words, given a measured increase in F in a test collection, a high UIR value makes more likely that an increase will also be observed in other test collections. Remarkably, UIR estimates cross-collection robustness of F increases much better than the absolute value of the F increase.

A limitation of our present study is that we have only tested UIR on the text clustering problem. While its usefulness for clustering problems already makes UIR a useful analysis tool, its potential goes well beyond this particular problem. Most Natural Language problems – and, in general, many problems in Artificial Intelligence – are evaluated in terms of many individual measures which are not trivial to combine. UIR should be a powerful tool in many of those scenarios.

An UIR evaluation package is available for download at http://nlp.uned.es.

## Acknowledgments

This research has been partially supported by the Spanish Government (grant Holopedia, TIN2010-21128-C02) and the Regional Government of Madrid under the Research Network MA2VICMR (S2009/TIC-1542).